\documentclass[fleqn,10pt,twocolumn]{SICE_FES25}

\usepackage{fancyhdr}
\usepackage{xcolor}

\title{Beyond Single Models: Enhancing LLM Detection of Ambiguity in Requests through Debate}

\author{Ana Davila${}^{1\dagger}$, Jacinto Colan${}^{2}$ and Yasuhisa Hasegawa${}^{1}$}
\speaker{Ana Davila}

\affils{${}^{1}$Institutes of Innovation for Future Society, Nagoya University, Nagoya, Aichi, Japan\\
${}^{2}$Department of Micro-Nano Mechanical Science and Engineering, Nagoya University, Nagoya, Aichi, Japan\\
}
\abstract{%
Large Language Models (LLMs) have demonstrated significant capabilities in understanding and generating human language, contributing to more natural interactions with complex systems. However, they face challenges such as ambiguity in user requests processed by LLMs. To address these challenges, this paper introduces and evaluates a multi-agent debate framework designed to enhance detection and resolution capabilities beyond single models. The framework consists of three LLM architectures (Llama3-8B, Gemma2-9B, and Mistral-7B variants) and a dataset with diverse ambiguities. The debate framework markedly enhanced the performance of Llama3-8B and Mistral-7B variants over their individual baselines, with Mistral-7B-led debates achieving a notable 76.7\% success rate and proving particularly effective for complex ambiguities and efficient consensus. While acknowledging varying model responses to collaborative strategies, these findings underscore the debate framework's value as a targeted method for augmenting LLM capabilities. This work offers important insights for developing more robust and adaptive language understanding systems by showing how structured debates can lead to improved clarity in interactive systems.}

\keywords{%
large language models, ambiguity, debate, human-robot interaction
}

\fancypagestyle{firstpage}{
  \fancyhf{}
  
  \fancyhead[L]{
    \vspace*{-25pt} 
    \parbox{\textwidth}{
      \textcolor{blue}{\small This work has been accepted at the 2025 SICE Festival with Annual Conference (SICE FES) and submitted to the IEEE for possible publication. Copyright may be transferred without notice, after which this version may no longer be accessible.}
    }
  }
}

\begin{document}

\maketitle
\thispagestyle{firstpage}


\section{Introduction}

Large Language Models (LLMs) have demonstrated significant capabilities in understanding and generating human language, contributing to more natural interactions with complex systems. However, a persistent challenge lies in the inherent ambiguity of human requests. Such ambiguities, if misinterpreted by an LLM, can lead to incorrect actions or responses, a notable issue in applications that require precise execution, such as task-oriented dialogue systems or human-robot collaboration \cite{ref_colan2025_review}. Even advanced single LLM architectures may not consistently identify and resolve these subtle but important sources of misunderstanding.

To address these limitations, this paper explores an approach that goes beyond single models. We propose and investigate a multi-agent debate framework where multiple LLMs collaboratively analyze a user's request. The rationale is that a structured debate, taking advantage of diverse perspectives from different LLM agents, may more effectively surface and scrutinize potential ambiguities than a single-model approach. This collaborative reasoning process is intended to enhance the detection of unclear elements within requests and facilitate the generation of targeted clarification questions, with the goal of leading to more consistent system behavior.

\begin{figure}[t]
\begin{center}
\includegraphics[width=0.48\textwidth]{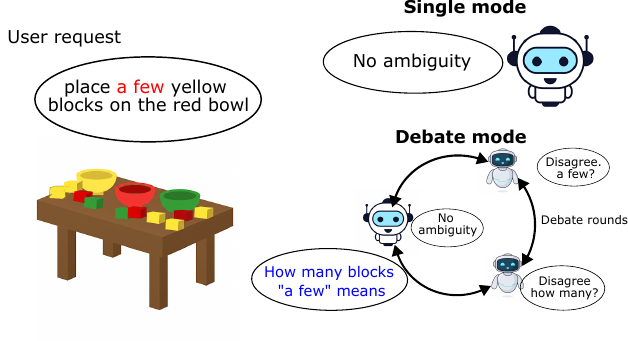}
\caption{\label{fig:1} Multi-agent debate flow}
\end{center}
\end{figure}

This study empirically evaluates the efficacy of such an LLM debate framework in improving the detection and resolution of ambiguities (Figure~\ref{fig:1}). We utilized a programmatically generated dataset that features various types of ambiguity commonly encountered in instructional contexts and tested the framework with three distinct LLM architectures. Our contribution lies in examining how this debate mechanism performs, revealing its effects on performance for certain models and ambiguity types, and identifying model-specific conditions under which it appears most advantageous. The findings provide insights into the conditional utility of collaborative LLM systems, which informs future efforts to develop a more reliable language understanding in interactive systems.

\section{Related works}

Resolving ambiguity in natural language commands is crucial for effective human-robot interaction (HRI), as vague instructions can lead to task failures \cite{ref_ichter2023,ref_ramachandran2025,ref_davila2024_voice}. Foundational HRI research addressed this, with early work on specific ambiguities. Liu et al. studied spatial ambiguity arising from differing frames of reference (e.g., ``to the right of the chair") \cite{ref_liu2010}, relevant to our study. Others developed systematic, non-LLM methods for clarification questions to resolve referential ambiguities (e.g., which ``medical kit") \cite{ref_williams2019} or attribute-based disambiguation for ``bring-me" tasks (e.g., object color) \cite{ref_morohashi2019}. Although effective for defined cases, these early systems had limited generalization. Our programmatic generation of data sets for numerical, attribute, and spatial ambiguities allows controlled investigation, complementing studies with general interaction data \cite{ref_chisari2025} and efforts to automate the generation of ambiguous scenarios \cite{ref_ichter2023}. This early work underscores the need for a robust clarification. Resolving such linguistic ambiguities is a key component of the larger challenge of recognizing the overall intent of the user and the task phase in real time \cite{ref_yamada2024_workflow}, and our study addresses this by focusing on optimal clarification questions.

Large Language Models (LLMs) have significantly advanced robot comprehension of complex, underspecified instructions, and implicit knowledge \cite{ref_singh2023,ref_chen2025}. LLMs are used for grounding instructions, planning, and, crucially, identifying ambiguities and formulating clarifying questions. Chisari et al. introduced AmbResVLM, grounding language goals in visual scenes for task ambiguity reasoning and query generation with an automated pipeline for ambiguous image-task datasets \cite{ref_chisari2025}. The ``Ask-to-Act" paradigm tasks embodied agents with asking minimal clarification questions for ambiguities in object attributes, spatial relationships, or size, using LLM reasoning \cite{ref_ramachandran2025}. Zhang et al. explored training LLMs with Reinforcement Learning from Human Feedback (RLHF) by simulating future turns, to learn what clarifying questions to ask and when \cite{ref_zhang2025}. Vision-Language Models (VLMs) like FreeGrasp resolve object ambiguities in cluttered scenes \cite{ref_freegrasp_anon2025} and enhance spatial reasoning \cite{ref_han2025}. Although LLMs allow for broader understanding, challenges such as grounding and hallucinations persist \cite{ref_singh2023,ref_feng2025}, making clarifications generated by LLM vital. Our study's use of smaller, locally deployable LLMs targets efficient, low-latency inference for real-time robotics.

To address single LLM limitations such as biases or presupposing one interpretation \cite{ref_lakara2025,ref_koupaee2025}, multi-LLM collaboration and debate are emerging \cite{ref_davila2025_ambiguity}. These frameworks improve reasoning and ambiguity resolution by allowing agents to critique and refine interpretations \cite{ref_lakara2025}. Lakara et al. proposed the LLM-Consensus, where agents assess contextual consistency, identify ambiguities, and refine reasoning, potentially using external information \cite{ref_lakara2025}. Koupaee et al.'s Madisse framework assigns initial stances to LLM agents for multi-round debates on summary faithfulness, uncovering errors and identifying ``ambiguity" from multiple correct interpretations \cite{ref_koupaee2025}. Such debates show potential for detecting subtle numerical, attribute, and spatial ambiguities. Hierarchical structures, such as leader-follower models, are also explored; ReMA uses a high-level metathinking agent for oversight and a low-level agent for execution \cite{ref_rema_anon2025}. 

The principles of multi-agent debate influence our leader-follower methodology by facilitating structured debate. This helps in considering multiple interpretations and evaluating clarifications for optimal questions, leveraging 'cognitive diversity' for effective ambiguity resolution. While frameworks like Madisse \cite{ref_koupaee2025} demonstrate debate for evaluating summary faithfulness, our work specifically applies and assesses a debate protocol for the pre-execution analysis of commands in a human-robot interaction context, focusing on generating actionable clarification questions.

\section{Methodology}

Our study investigates the effectiveness of single and multi-agent Large Language Model (LLM) configurations in identifying ambiguities within user instructions and proposing pertinent clarifying questions. 

The methodology encompasses a programmatically generated ambiguity dataset, a selection of LLM agents chosen for their suitability for such scenarios, and a structured experimental protocol involving debate mechanisms.

\subsection{Ambiguity Dataset Generation}
The experiments used a custom dataset of ambiguous instructions, programmatically generated to ensure controlled variation and verifiability. The generation process was automated to create pairs of ambiguous and unambiguous instructions, each grounded in a consistent, predefined scenario context describing a simple environment with a fixed set of objects (e.g., three blocks each of red, yellow, and green, and one bowl of each color on a table). Each entry in this dataset is stored in a structured data format (JSON), including a unique identifier, the shared scenario context, the ambiguous instruction, its corresponding unambiguous counterpart, and the type of ambiguity introduced.

The generation process aimed at three primary types of ambiguity:

\begin{enumerate}
    \item \textbf{Numerical Ambiguity}: These instances focused on vague quantities of objects, specifically blocks. An unambiguous instruction would specify a precise quantity (e.g. ``two red blocks," or ``all green blocks" that resolved to three). The corresponding ambiguous instruction was created by replacing the precise quantity with an imprecise term (e.g., ``a few red blocks," ``some green blocks") while keeping other elements of the command—such as action, object color, spatial relationship, and target object—identical to its unambiguous counterpart. Actions (e.g., ``put," ``place," ``move"), object attributes, and relational components were selected from predefined lists.

    \item \textbf{Attribute Ambiguity}: This category involved ambiguity in specifying the properties of an object, either its type (noun) or its color. An unambiguous instruction would use precise terms (e.g., ``a single red block"). For noun-based ambiguity, the ambiguous version would replace the specific object noun (e.g., ``block") with a more general or related term (e.g., ``cube," ``item," ``thing") while retaining the correct color. For color-based ambiguity, the specific color term (e.g., ``red") was replaced with a less common synonym or a descriptive phrase (e.g., ``crimson," ``cherry-colored"), while the object noun remained correct. The rest of the sentence structure, including quantity (typically singular for attribute tests), action, and relational components, was maintained.

    \item \textbf{Spatial Ambiguity}: These cases focused on imprecise descriptions of spatial relationships between objects. An unambiguous instruction would use a clear and specific spatial preposition (e.g., ``on the yellow bowl," ``to the left of the green block"). The ambiguous instruction was generated by substituting the precise preposition with a more vague or underspecified one (e.g., ``near the yellow bowl," ``lateral to the green block," ``along the line of sight of the red block"). The objects being manipulated and referenced, along with the action, were kept consistent between the ambiguous and unambiguous pair.
\end{enumerate}
For all types, the generation procedure randomly selected components from predefined vocabularies to construct a specified number of unique examples for each ambiguity category. Care was taken to ensure that the referenced objects were distinct from the primary manipulated object, when appropriate. The resulting dataset provided a controlled yet varied set of challenges for the LLMs.

\subsection{LLM Agents}
Given the intended application in real-time human-robot collaboration, the selection of LLM agents prioritized models with the potential for efficient local deployment and low-latency inference. This is crucial for interactive scenarios where quick responses are necessary for a natural and effective collaborative experience. Therefore, we opted for relatively smaller, yet capable models. In this study, three distinct LLMs were used as debaters:

\begin{itemize}
    \item Llama-3-8b-instruct
    \item Gemma-2-9b-it
    \item Mistral-7b-instruct
\end{itemize}

These models showcase a variety of open-source architectures known for their popularity and high performance. This diversity enables exploration of how different reasoning patterns interact in a collaborative setting. They offer a balance between high performance and manageable computational demands, making them suitable for deployment on local GPUs commonly found in robotic environments.
Each model instance, referred to as an ``agent," was configured with consistent generation parameters, such as a fixed temperature (0.5) to control randomness and a maximum token limit (350 tokens) for responses, ensuring comparability.

\subsection{Experimental Procedure}
The core task of the LLMs was to analyze a given ambiguous instruction within its specified scenario context. The objective was to determine whether the command is sufficiently clear for execution or, if deemed ambiguous, to propose a single optimal clarifying question. Agents were instructed to limit their reasoning to a predefined maximum number of sentences (e.g., four sentences). We implemented two primary experimental conditions:

\subsubsection{Single-Agent Baseline}
In this configuration, each of the three LLM agents independently processed each ambiguous instruction from the dataset. The agent received the scenario context and the ambiguous instruction and was prompted to provide its reasoning (within the sentence limit) and then concluded its response by stating that the instruction is clear or by proposing a single, best clarifying question if ambiguity was detected. This was a single-round interaction, and the agent's output (argument and proposal) along with processing latency was recorded.

\subsubsection{Multi-Agent Debate (Leader-Follower Protocol)}
This condition involved a structured debate between three agents: one designated as the ``Leader" and the other two as ``Followers". A two-follower configuration was chosen to establish a stronger consensus mechanism, requiring the leader to convince two independent agents, thereby reducing the risk of premature agreement on a flawed interpretation. The roles were rotated among the three LLM models to ensure that each model acted as a leader for each ambiguous instruction, thus mitigating potential biases associated with a fixed leader. The debate proceeded in rounds, with a predefined maximum number of rounds (e.g., five rounds).

\begin{description}
    \item[Round 1: Leader's Initial Proposal] The Leader agent received the scenario context and ambiguous instruction, then generated an initial proposal consisting of its reasoning and either a verdict of clarity or a specific clarifying question.

    \item[Round 1: Follower Evaluation] Each Follower agent received the original context, instruction, and Leader's proposal. Followers were prompted to state their stance (``Agree" or ``Disagree"), provide reasoning, and if disagreeing, they could offer an alternative question. Specifically, if the Leader proposed a question, a disagreeing Follower could provide a different one; if the Leader declared clarity, a disagreeing Follower's reasoning was expected to explain the ambiguity, optionally proposing a question or indicating no alternative was needed. Agreeing Followers indicated no alternative question.

    \item[Consensus Check] If both Followers agreed with the Leader's proposal from Round 1, consensus was considered reached, and the debate for that instruction concluded with the Leader's proposal as the final outcome.

    \item[Subsequent Rounds (up to the maximum limit)] If consensus was not reached, the Leader received all follower feedback. It then synthesized this feedback to generate a new proposal (reasoning and then either a verdict of clarity or a new/revised clarifying question). This new proposal was then presented to the Followers for evaluation under the same protocol. A consensus check followed each follower evaluation phase.

    \item[Termination Conditions] The debate for a given instruction terminated if:
    \begin{itemize}
        \item Consensus was achieved (both Followers agreed with the Leader's proposal).
        \item The maximum number of configured rounds was completed without consensus. In this scenario, the outcome was marked as non-consensus, and the final state of proposals and feedback was recorded.
        \item An unrecoverable error occurred during an agent's response generation or in parsing the structured output.
    \end{itemize}
\end{description}
The experimental procedure was applied to individual entries from the ambiguity dataset, ensuring that for each entry, all single-agent baselines were run, and for the multi-agent condition, all rotations of the leader role were executed.

\section{Experimental Results}
\label{sec:results}

This section analyzes the performance of single and multi-agent LLM configurations in ambiguity detection, focusing on key trends and insights derived from the experimental data. The figures mentioned are assumed to graphically represent the specific data values previously detailed.

\subsection{Overall Performance and Model Dependencies}

\begin{figure}[h]
\begin{center}
\includegraphics[width=0.48\textwidth]{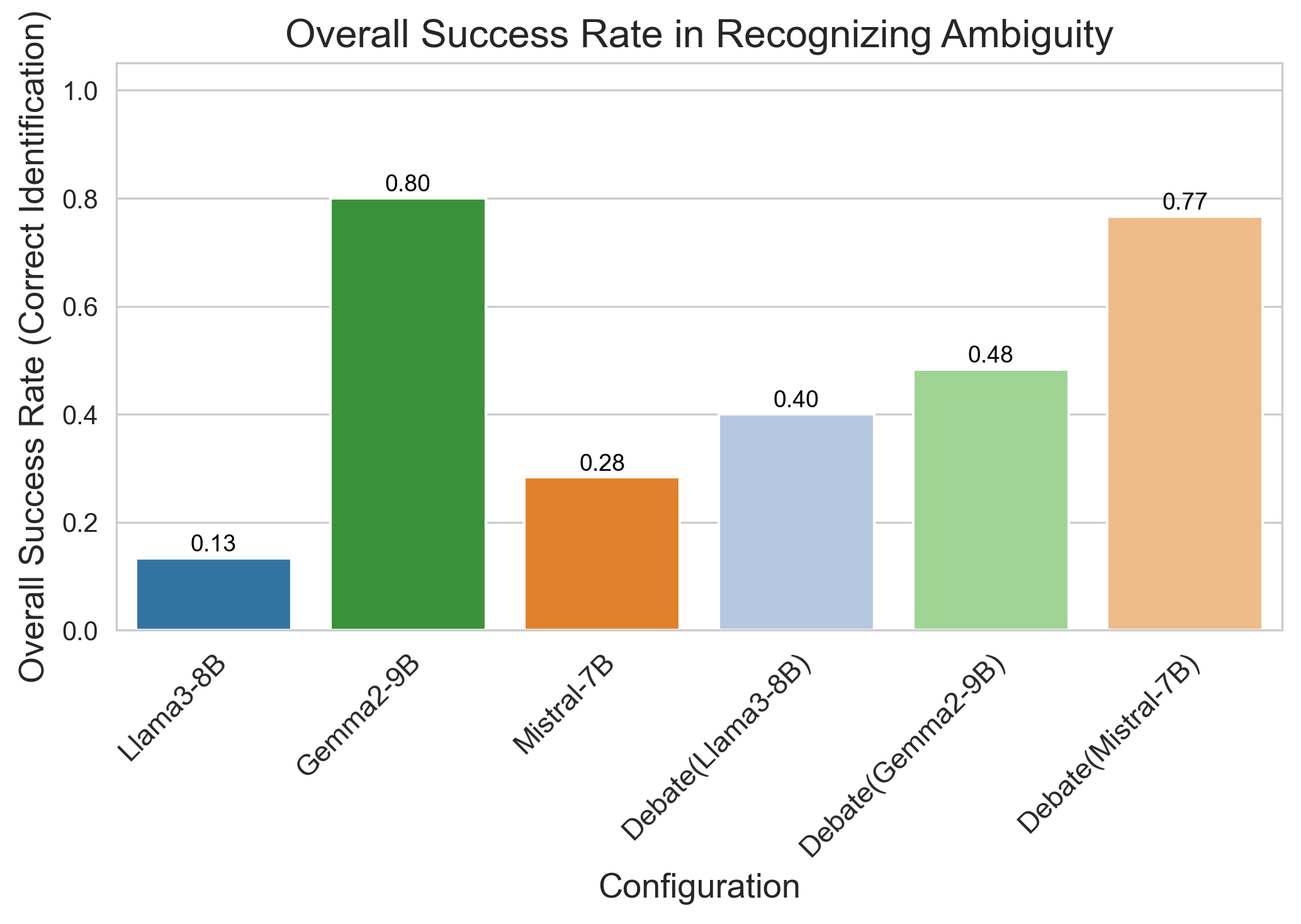}
\caption{\label{fig:overall_success} Overall success rate in recognizing ambiguity for single LLM agents (Llama3-8B, Gemma2-9B, Mistral-7B) and multi-agent debate configurations.}
\end{center}
\end{figure}

Figure \ref{fig:overall_success} shows that the single Gemma2-9B agent achieved a success rate of $80.0\%$. This rate was considerably higher than that of single Llama3-8B ($13.3\%$) and Mistral-7B ($28.3\%$).

The multi-agent debate framework yielded varied outcomes. For Mistral-7B, debates improved performance from $28.3\%$ to $76.7\%$ (Debate(Mistral-7B)), resulting in the highest success rate among the debate configurations. Llama3-8B also saw an increase in performance, from $13.3\%$ to $40.0\%$ with debate. In contrast, the setting of the debate appeared detrimental to Gemma2-9B, as Debate(Gemma2-9B) achieved success $48.3\%$, which was less than its single agent performance. This suggests that the utility of a debate structure can be model-dependent and may not universally guarantee improvement. For individual models like Gemma2-9B that demonstrate high baseline performance, the current debate mechanism might interfere with their individual processing capabilities.

\subsection{Impact of Ambiguity Type}

\begin{figure}[h]
\begin{center}
\includegraphics[width=0.48\textwidth]{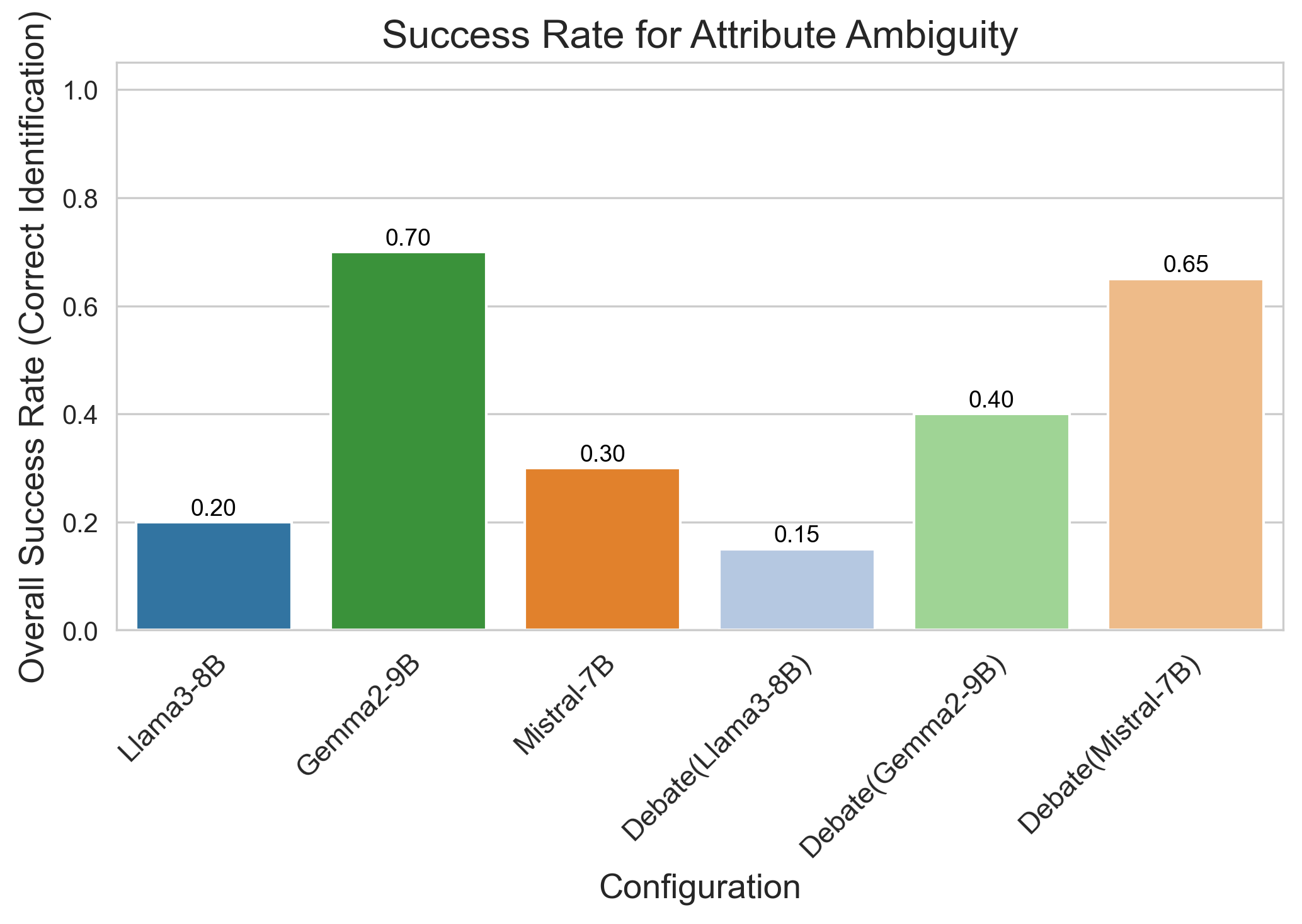}
\caption{\label{fig:attribute_success} Success rate for Attribute Ambiguity across different LLM configurations.}
\end{center}
\end{figure}

\begin{figure}[h]
\begin{center}
\includegraphics[width=0.48\textwidth]{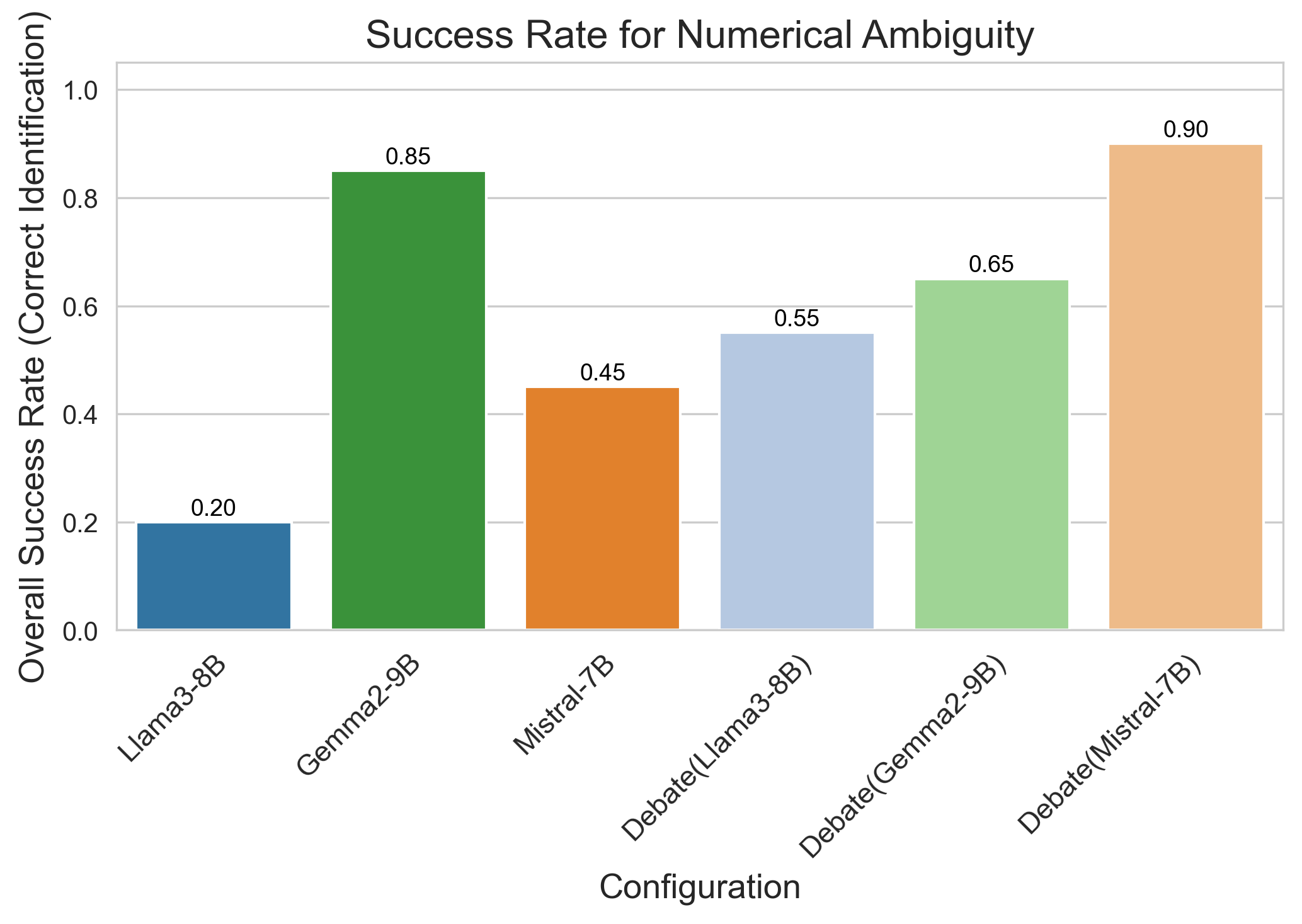}
\caption{\label{fig:numerical_success} Success rate for Numerical Ambiguity across different LLM configurations.}
\end{center}
\end{figure}

\begin{figure}[h]
\begin{center}
\includegraphics[width=0.48\textwidth]{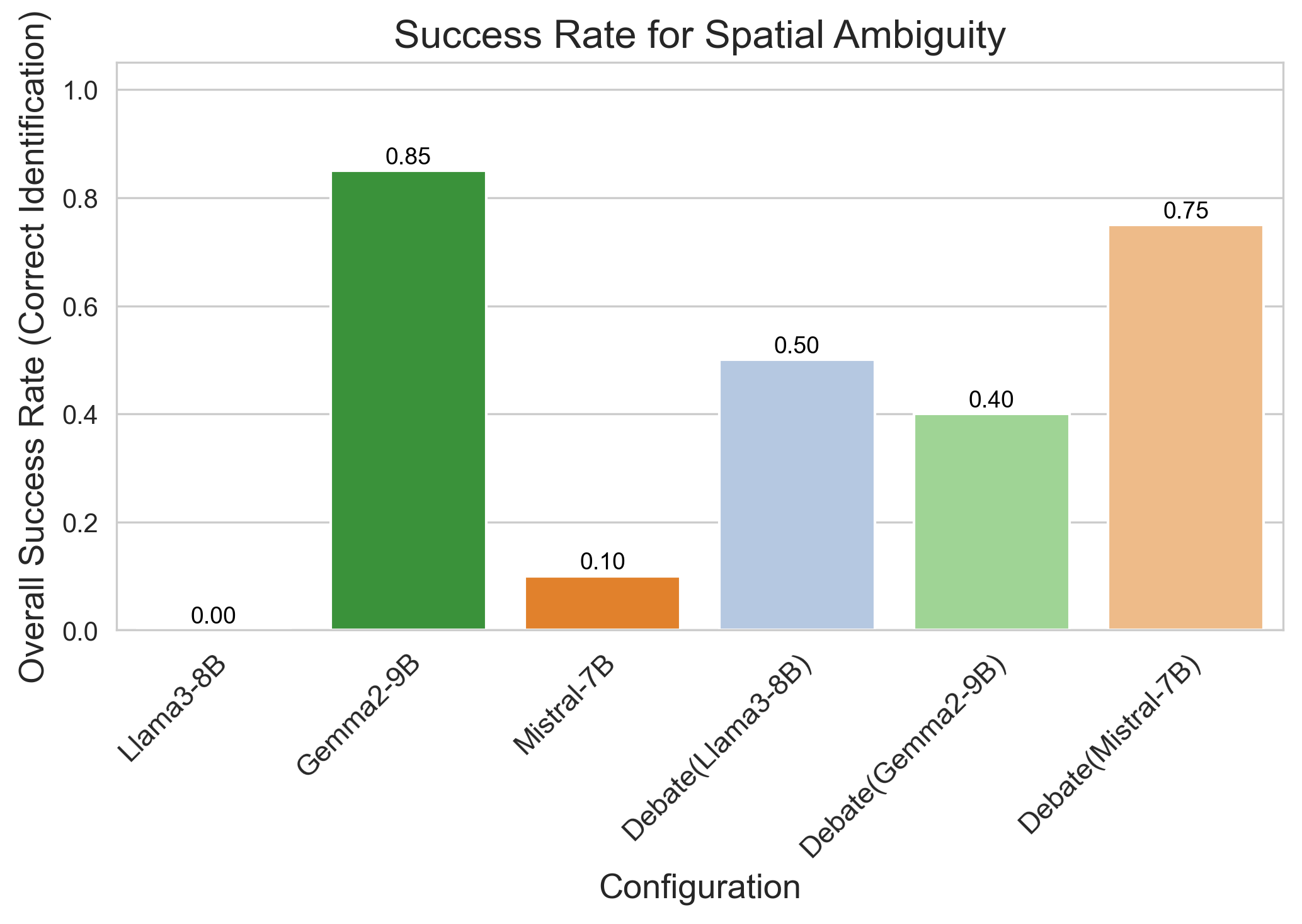}
\caption{\label{fig:spatial_success} Success rate for Spatial Ambiguity across different LLM configurations.}
\end{center}
\end{figure}

Performance varied significantly by type of ambiguity, as detailed in Figures \ref{fig:attribute_success}, \ref{fig:numerical_success}, and \ref{fig:spatial_success}.
Single Gemma2-9B performed consistently well across attribute ($70\%$), numerical ($85\%$), and spatial ($85\%$) ambiguities. In contrast, Llama3-8B and Mistral-7B had a lower performance as single agents, particularly with spatial ambiguities (Llama3-8B: $0\%$, Mistral-7B: $10\%$).

The debate configuration led by Mistral-7B showed notable improvements over its single-agent counterpart, particularly for more complex types, achieving $65\%$ in attribute, $90\%$ in numerical, and $75\%$ in spatial ambiguities. For instance, the increase in spatial ambiguity detection for Mistral-7B from $10\%$ (single) to $75\%$ (debate) illustrates this effect. However, the Debate(Mistral-7B) configuration did not consistently achieve higher success rates than the single Gemma2-9B across all types, with the single Gemma2-9B showing higher rates for attribute and spatial tasks. Some debate configurations also performed lower than their single-agent counterparts on specific tasks (e.g., Debate(Llama3-8B) on attribute ambiguity).

\subsection{Debate Dynamics: Consensus and Efficiency}

\begin{figure}[h]
\begin{center}
\includegraphics[width=0.48\textwidth]{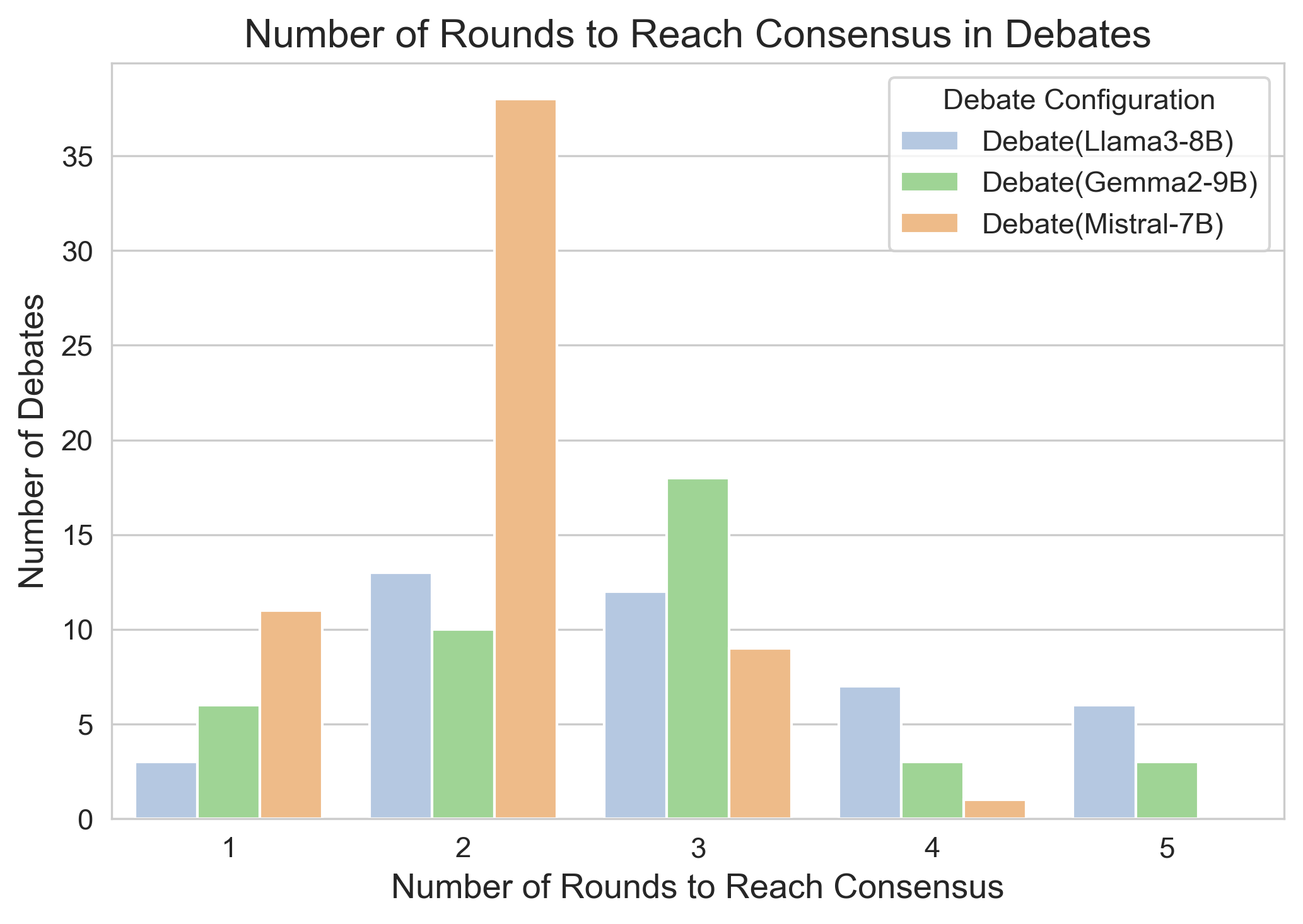}
\caption{\label{fig:rounds_to_consensus} Distribution of the number of rounds required to reach consensus in multi-agent debates that achieved consensus.}
\end{center}
\end{figure}

\begin{figure}[h]
\begin{center}
\includegraphics[width=0.48\textwidth]{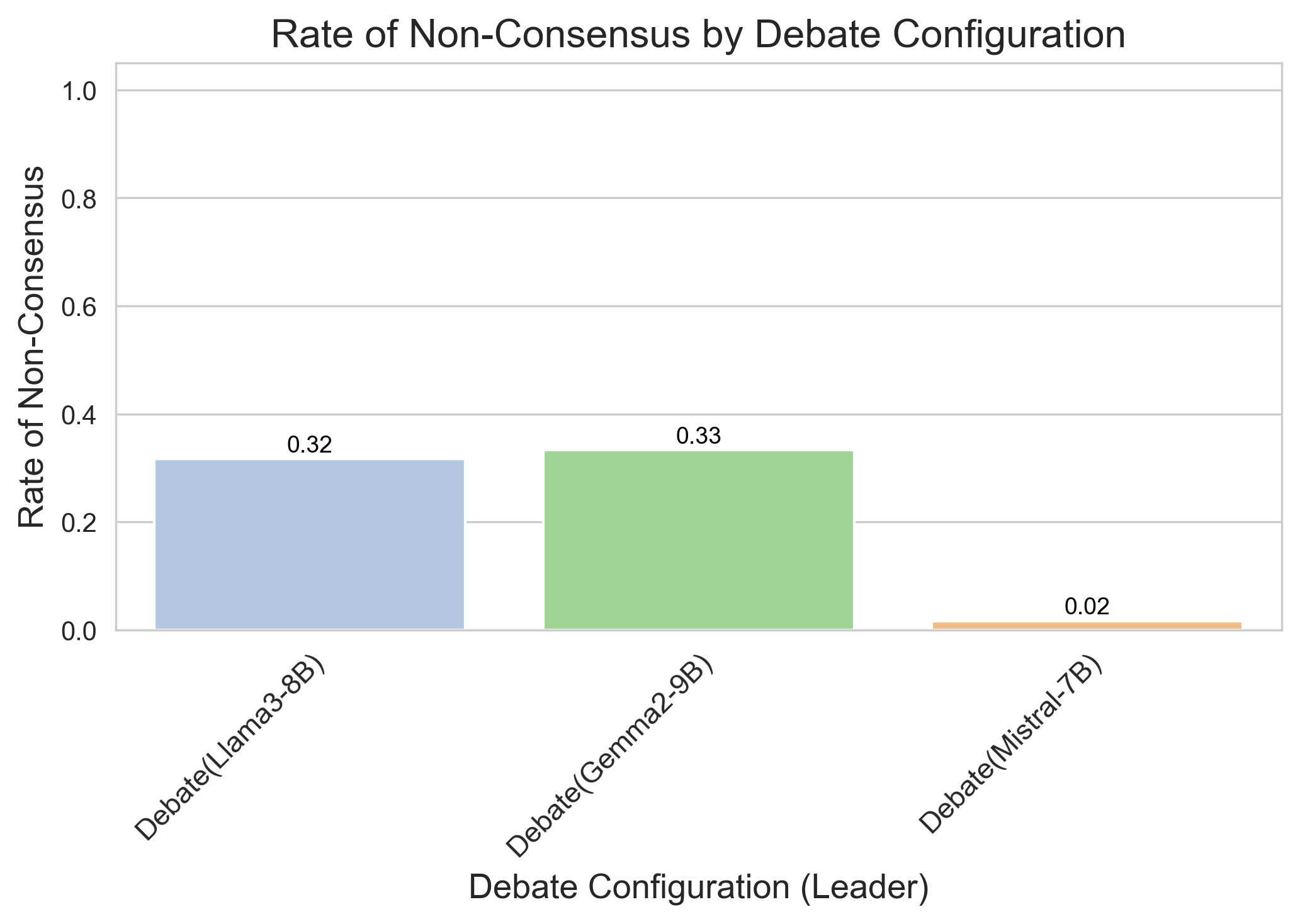}
\caption{\label{fig:non_consensus_rate} Rate of non-consensus by debate configuration.}
\end{center}
\end{figure}

The dynamics of the debate process (Figures \ref{fig:rounds_to_consensus}, \ref{fig:non_consensus_rate}) revealed further differences. The debates led by Mistral-7B achieved consensus with high frequency ($98.3\%$ consensus rate, corresponding to a nonconsensus rate of $1.7\%$), predominantly within one or two rounds (with $64.4\%$ of its consensus occurring in Round 2). This suggests effective proposal generation and feedback integration when Mistral-7B acted as a leader.

In contrast, the debates led by Llama3-8B and Gemma2-9B had considerably higher non-consensus rates (approximately $31.7\%$ and $33.3\%$, respectively). These configurations also tended to require more rounds to reach consensus when it was achieved; for instance, $45\%$ of Gemma2-9B's successful consensuses occurred in Round 3. This may indicate a greater difficulty in reconciling differing views or different synthesis capabilities when these models led debates.

The data in Table \ref{tab:leader_effectiveness} indicate that when consensus was reached in those particular experimental runs, the decisions were generally of high quality (e.g., $90-95\%$ success for Mistral-7B and Llama3-8B led debates, as presented in Table \ref{tab:leader_effectiveness}). However, data from the plots related to Figures \ref{fig:rounds_to_consensus} and \ref{fig:non_consensus_rate} suggest that achieving consensus efficiently and consistently varied between configurations.

\begin{table}[htb]
\caption{\label{tab:leader_effectiveness}Leader LLM Effectiveness in Debates. Metrics include percentage reaching consensus, average rounds to consensus, and success rate of consensus outcomes.}
\begin{center}
\begin{tabular}{|l|c|c|c|}
\hline
\textbf{Leader} & \textbf{Cons. } & \textbf{Avg.} & \textbf{Success} \\
& \textbf{Reach (\%)} & \textbf{Rounds} & \textbf{(\%)} \\
\hline
Llama3-8B   & 85 & 1.8 & 92 \\ \hline
Gemma2-9B       & 82 & 2.1 & 88 \\ \hline
Mistral-7B        & 90 & 1.5 & 95 \\ \hline
\end{tabular}
\end{center}
\end{table}

\subsection{Temporal Costs}

The average time for the debates, shown in Table \ref{tab:avg_debate_time}, ranged from $22.5$ to $28.1$ seconds. Debates led by Mistral-7B were the quickest in this set of data, which aligns with their tendency to reach consensus efficiently. This temporal overhead is a consideration for real-time applications. For models like Llama3-8B and Mistral-7B (based on plot analyses), the observed accuracy gains from debate might be weighed against this cost. For Gemma2-9B, its high single-agent performance, presumably achieved at lower latency, suggests that the added time and reduced accuracy of debates in this study represent a less favorable trade-off.

\begin{table}[htb]
\caption{\label{tab:avg_debate_time}Average debate time (seconds) by leader LLM.}
\begin{center}
\small 
\begin{tabular}{|l|c|}
\hline
\textbf{Leader LLM} & \textbf{Avg. Debate Time (s)} \\
\hline
Llama3-8B Instruct  & 25.3 \\ \hline
Gemma2-9B           & 28.1 \\ \hline
Mistral-7B          & 22.5 \\ \hline
\end{tabular}
\end{center}
\end{table}

\section{Conclusion}
\label{sec:conclusion}

To address challenges single Large Language Models (LLMs) face with ambiguous requests in robotics, this study demonstrated the value of a multi-agent debate framework. Our investigation revealed that collaboration significantly improves performance for models such as Llama3-8B and Mistral-7B. The debates led by Mistral-7B were particularly successful, achieving a success rate of 76. 7\% and proved to be effective in resolving complex spatial ambiguities and efficiently building consensus. However, the utility of the framework is model-dependent. It degraded the performance of the single agent with high performance, Gemma2-9B (80. 0\% success). These findings affirm the potential of the debate framework as a targeted enhancement tool while underscoring that it is not a universally optimal solution.

The model-dependent outcomes suggest that the effectiveness of the debate is based on the complementary capabilities of the agents. For some models, structured feedback helps overcome individual reasoning gaps by forcing the consideration of alternative interpretations. Conversely, for an already strong agent like Gemma2-9B, the debate may introduce distracting noise from other agents, steering it from a correct initial assessment. This insight marks a step toward more robust and adaptive systems in Human-Robot Interaction (HRI). Recognizing the current temporal overhead (22.5--28.1s) as a challenge for real-time use, future work will focus on optimizing the protocol, for example, with adaptive triggers that invoke debate selectively, and conducting deeper qualitative analyses of failure modes and prompt sensitivity.


\end{document}